\documentclass{article}
\usepackage{spconf,amsmath,graphicx}
\usepackage{amsfonts,setspace}
\usepackage{amssymb}
\usepackage{graphicx}
\usepackage{pifont}% http://ctan.org/pkg/pifont
\usepackage{hyperref}
\usepackage{xcolor}
\title{Privacy-Preserving Deep Learning Using Deformable Operators for Secure Task Learning \vspace{-1mm}}
\name{Fabian Perez \qquad Jhon Lopez \qquad Henry Arguello \thanks{ code is available at: \url{https://github.com/Factral/PrivDL}}}
\address{ Department of Computer Science, Universidad Industrial de Santander\\ Bucaramanga, 680002, Colombia \vspace{-3mm}}% \\
\begin{document}

%\ninept
%

\maketitle
\begin{abstract}

In the era of cloud computing and data-driven applications, it is crucial to protect sensitive information to maintain data privacy, ensuring truly reliable systems. As a result, preserving privacy in deep learning systems has become a critical concern. Existing methods for privacy preservation rely on image encryption or perceptual transformation approaches. However, they often suffer from reduced task performance and high computational costs. To address these challenges, we propose a novel Privacy-Preserving framework that uses a set of deformable operators for secure task learning. Our method involves shuffling pixels during the analog-to-digital conversion process to generate visually protected data. Those are then fed into a well-known network enhanced with deformable operators. Using our approach, users can achieve equivalent performance to original images without additional training using a secret key. Moreover, our method enables access control against unauthorized users. Experimental results demonstrate the efficacy of our approach, showcasing its potential in cloud-based scenarios and privacy-sensitive applications.

\end{abstract}

\begin{figure*}[htb]

\begin{minipage}[b]{1.0\linewidth}
  \centering
  \centerline{\includegraphics[width=17cm]{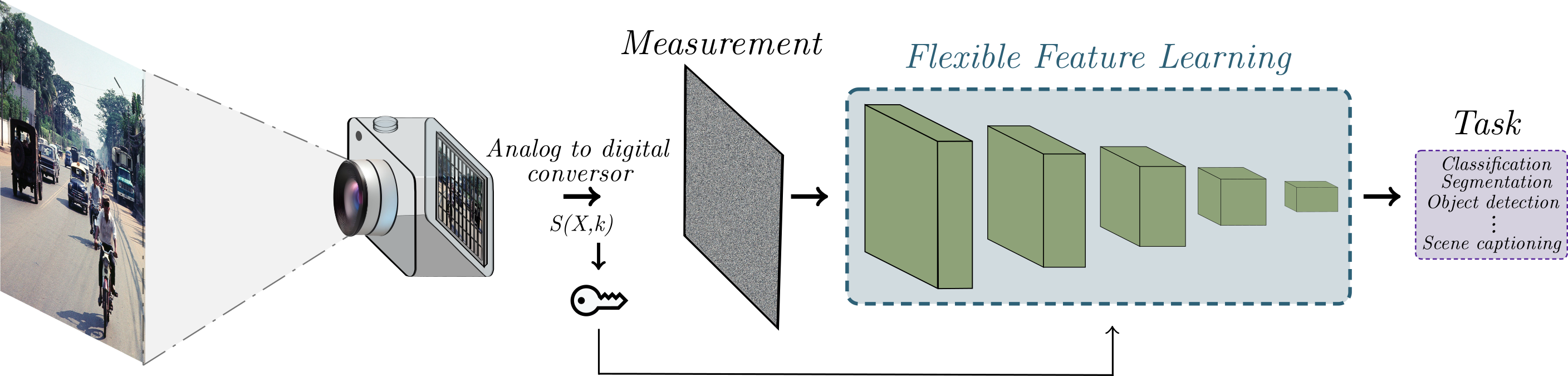}}
%  \vspace{2.0cm}
  \caption{Proposed framework for flexible feature learning from private images. The camera captures an image, which is then passed through a custom analog-to-digital converter to apply a transformation. The resulting measurement is a private image that is an input to the flexible feature learning module. This module generates underlying features that can be used in any task such as classification, object detection, and segmentation.}\medskip
\end{minipage}

\end{figure*}

\section{Introduction}
\label{sec:intro}

As deep neural networks continue to be deployed in real-world applications, mainly as software-as-a-service tools used through cloud computing platforms, users routinely upload their sensitive data. For instance, medical records or images containing private information to third-party servers, to mention a few. However, this exposes users to potentially serious privacy risks through unauthorized data access or information leakage \cite{tanuwidjaja2020privacy, huang2014survey}. Since users have little control over how their information is used once uploaded. Thus, this lack of control raises significant privacy and security concerns. Consequently, it demands privacy-preserving deep learning techniques to ensure reliable and secure systems, as urged increasingly in the industry.

To address these privacy and security issues, various approaches have been proposed to protect sensitive data in deep learning systems. These approaches can be broadly categorized into two main frameworks: fully homomorphic encryption (FHE) \cite{disabato2020privacy, kim2023optimized} and perception-based transformations (PT) \cite{kiya2022overview}. FHE methods enable operations on encrypted data and models equivalent to computations in the unencrypted domain. However, these techniques are restricted to shallow and narrow models and simple tasks due to the high computational cost of FHE operations. This makes them impractical for most deep learning architectures, which require more complex models and tasks.
On the other hand, PT techniques directly modify the input data to obscure sensitive information. Traditional approaches such as learnable encryption (LE) \cite{LE} and their extended version (ELE) \cite{madono2020block} apply transformations such as block-wise scrambling, pixel shuffling, negative-positive flipping, and channel shuffling to encrypt images visually. Nonetheless, they require additional adaptation networks to enable the encrypted images to be compatible with classification models, which implies significant computational overhead. Other works have utilized different backbones like convMixer \cite{qi2023privacy} or vision transformers \cite{qi2022privacy} for the adaptation network. However, while these methods explore alternative network architectures, they need to address the fundamental issue of additional computational costs sufficiently. Furthermore, other methods learn the transformation instead of making it explicit by using Autoencoders \cite{ito2021image} or GANs \cite{sirichotedumrong2021gan} to generate images very different from the original. 

Moving beyond these techniques, another line of work within PT is to design optics and algorithms for learning the ad hoc transformation \cite{sitzmann2018end}. For example, recent works have proposed designing a refractive camera lens to generate blurred images while extracting the most valuable features for the task through end-to-end optimization \cite{arguello2022optics}. This approach solves the problem of capturing the original image and the computational cost but highlights the issue of the inherent privacy-utility trade-off. All methods mentioned so far exhibit this trade-off, where greater privacy comes at the cost of reduced primary task accuracy. A recent work, Key-Nets \cite{byrne2020key}, progresses in this direction by designing a custom sensor that applies optical and analog transformations reflected in multiple keys. Those are then used in a convolutional network modeled with sparse Toeplitz matrices for creating keyed layers that can perform exact encrypted inference with this model. They can eliminate the privacy-utility trade-off, but the approach must still fully resolve the problem, as Key-Nets only allow using pre-trained networks containing linear or ReLU layers.
Moreover, they add complexity by requiring custom optical components. In general, existing perceptual transformation techniques face limitations in balancing privacy-utility and computational costs, which restricts their application in real-world systems. Then, overcoming this problem remains an open challenge. Therefore, motivated to overcome these limitations, we introduce a novel privacy-preserving deep learning framework based on deformable operators \cite{dai2017deformable}. Specifically, our approach involves applying a secret spatial transformation to shuffle pixels during acquisition.
Consequently, private images are obtained and fed into well-established convolutional neural network (CNN) architecture enhanced with deformable convolutions and deformable max pooling. This allows our framework to perform equivalent tasks to original images without requiring additional training or multi-key management. Our method also enables access control for intellectual property protection by using the secret transformation as a key to be shared only with authorized users \cite{ito2021access}. Deformable operators allow our privacy-preserving approach to overcome the inherent privacy-utility trade-off and computational cost of existing techniques.

\section{Proposed Method}

Our proposed method consists of two main modules: the Perceptual Image Transformation, which is applied directly in the acquisition process, and the Flexible Feature Learning (FFL), which leverages deformable operators to extract meaningful features from the private data. These two modules are closely connected through the key generated in the first module. This enables the model to be adapted to private images, making it functional for the desired task.

\subsection{Perceptual Image Transformation}

We propose a perceptual image transformation module that applies spatial pixel shuffling to make images visually private. This pixel shuffling can be performed in the acquisition process during the analog-to-digital conversion. Hence, it allows the direct capture of private images without needing post-capture encryption. This transformation can be modeled as:

\begin{equation}\label{eq:1}
    X' = \mathcal{S}(X, K),
\end{equation}
\noindent
where $\mathbf{X} \in \mathbb{R}^{h \times w \times c}$ is the original image,  $\mathbf{K} \in \mathbb{Z}_{+}^{h \times w}$ is the key to stores the rearrangement information to obtain the privatized image. $\mathcal{S}(\cdot, \cdot)$ is the shuffle function parameterized by the key $\mathbf{K}$ where rearranges all the pixels of $\mathbf{X}$ to generate the private image $\mathbf{X'}$. In addition, h, w, and c represent height, width, and number of image channels. The operation of shuffling is applied at the level of pixels. Specifically, the pixel at position $(i,j)$ in $X'$ is taken from the pixel at position $k(i,j)$ in $X$. This process is repeatedly applied for each channel. This indexing scheme gives the key $k$ a seed for the shuffling, allowing for controlled private image generation. If this function $\mathcal{S}$ is applied during image acquisition with a custom analog-to-digital algorithm, we can directly obtain encrypted images $X'$ without needing post-processing. In this scenario, when the shuffling operation is applied, the key $k$ must be kept secret and stored using a traditional encryption method such as AES or RSA to prevent information leakage. Then, it can be applied in the flexible feature learning module to process the private images for a specific task.

\subsection{Flexible Feature Learning}

This FFL module enables processing the desired task of the private images generated by the perceptual transformation module while maintaining both task performance and privacy. This is achieved by leveraging deformable max pooling and deformable convolutions \cite{dai2017deformable}, where the latter, is a modified version of the convolutional operator. This deformable convolutional and max pooling operator employs additional offsets to adapt the sampling grid of the kernel dynamically. Hence, it allows free-form deformation of the sampling grid, which enables more flexible feature learning and greater robustness to spatial variations.

In our approach, we use deformable convolutional layers and initialize the offset $\Delta \mathbf{p}$ of the first deformable convolution layer with the key used in the transformation defined in Eq.\ref{eq:1}. This key shuffles the pixels during image acquisition and reflects the spatial shuffling applied during the perceptual transformation module (Sec 2.1). As a result, the deformable convolutions and max pooling can account for the pixel shuffling directly in the kernel sampling of the task operations, enabling compatible feature extraction from the private images. These features will be equivalent to the features of a plain image. This is achieved due to the offset $\Delta \mathbf{p} \in \mathbb{R}^{h' \times w'\times (2*n*n)}$, where $h'$ and $w'$ are the output height and width calculated from the deformable operator, and $n$ is the kernel size of the operator layer. In this representation, each pixel represents a sampling location in the operator grid, i.e., the offsets encode displacements in the $x$ and $y$ directions for each pixel in the sampling grid.

\begin{figure}[!h]
    \centering
    \includegraphics[width=7.5cm]{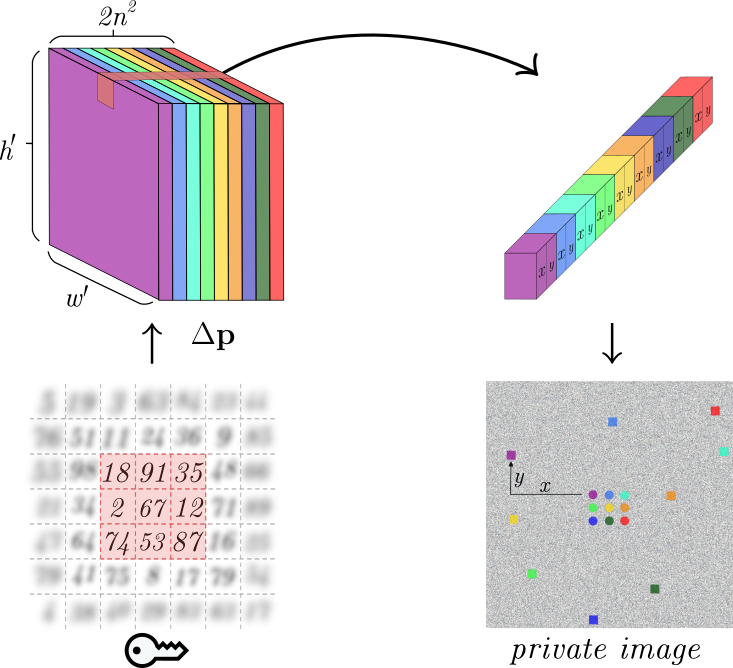}
    \caption{Offset generation from a key. The key stores the true pixel positions, while  $\Delta \mathbf{p}$ holds the distance to the true position from the kernel sampling; each pixel across the last dimension corresponds to a sampling of the deformable operator. }
    \label{fig:offset_generation}
\end{figure}

Leveraging this, the offset is constructed from the key $k$ to mimic convolutions and pooling on original images and obtain the same feature maps after the operation. A function achieves this
\begin{equation}
    f: \mathbb{R}^{h \times w} \to \mathbb{R}^{h' \times w' \times (2*n*n)},
\end{equation}

to map the key into offset values, with the offset constructed for a key, it is set in the first deformable convolution layer, as illustrated in Fig. \ref{fig:offset_generation}. To keep feature maps shuffled throughout the network and avoid revealing sensitive information during offset construction, displacements are taken from a randomly generated key $k_1(i,j)$ inside the network instead of $k(i,j)$, the $x$,$y$ displacements are calculated from $k$. This process is done for each pixel to fill the entire offset volume. For subsequent layers, the process is similar: random keys $k$ shuffle feature maps and offsets maintain compatibility with encryption, at the minimal latent space in layer $n$.  $k_n$ re-orders the feature map if necessary, which reveals no sensitive information due to it being sufficiently compressed and condensed at that stage. All this process allows pre-trained weights on plain images to be directly loaded for inference without re-training on the private images.

\begin{table}[!t]
\centering
\begin{tabular}{lc}
\hline
Experiment                           & Accuracy$(\%)$  \\ \hline
Non-privacy     & $\mathbf{94.1}$             \\
Pre-trained    & $11.7$              \\
Fine-tunning & $57.6$              \\
our method with key& $\mathbf{94.1}$               \\
our method without  key       & $11.8$              \\ \hline
\end{tabular}%
\caption{Privacy validation results for the proposed experiments. The best results are shown in boldface.}
\label{tab:validation_results}
\end{table}

\section{Results}

To evaluate the efficacy of our proposed framework for performing specific tasks, we conducted a series of experiments using image classification and semantic segmentation tasks. Specifically, we conducted a series of experiments on the CIFAR-10 dataset for classification. To achieve this, we utilized a VGG-16 \cite{simonyan2014very} architecture where all standard convolutions and max poolings were replaced by deformable convolutions and deformable max poolings to be compatible with the flexible feature learning module (FFL). For all experiments, we generated random keys for the module of perceptual transformation.

\noindent\textbf{Privacy Validation}. To validate the privacy protection of our approach, we evaluated the performance of the shuffled images for the classification task under the following experiments. \textit{\textbf{Non-privacy:}} we trained a VGG network on the original images without any privacy protection. \textit{\textbf{Pre-trained:}} we evaluated the network trained in the previous experiment by providing private images as input. \textit{\textbf{ Fine-tuning:}} we fine-tuned the VGG network on the private images. \textit{\textbf{Our method with correct key:}} we loaded the non-privacy network and replaced convolutions and poolings with deformable operators, the offset is fixed with the key, and the private images are evaluated in the network. \textit{\textbf{Our method with wrong key:}} we utilized the network from the previous experiment and evaluated it in the case where an incorrect key is fed

As shown in Table \ref{tab:validation_results}, our proposed method performs identically to the non-private method while maintaining privacy protection. When evaluating a network trained on plain images directly on the private images, the accuracy drops to almost random guess levels (11.7\%), showcasing the strong privacy guarantees of our perceptual transformation; fine-tuning the pre-trained model on the private images can recover some performance (57.6\%) by adapting to the protected images, but this reduces utility and requires additional training. In contrast, our approach matches the accuracy of the original non-private model (94.1\%) by using the correct key in the deformable convolutions, enabling precise feature extraction from the protected images without extra training. Moreover, when using an incorrect key, the performance again drops to near-random (11.8\%), demonstrating the access control capability of our technique; only with the correct key can the authorized user achieve full utility on the private data. Overall, these comprehensive experiments validate the efficacy of our proposed framework to deliver strong privacy protection with minimal impact on task performance for sensitive deep learning applications.

Once we demonstrated the efficacy of our proposed method under the outlined settings, we evaluated our performance of the proposed approach using a PreResNet-110 \cite{he2016identity} adapted to our method compared to state-of-the-art techniques for private image classification.
% \usepackage{graphicx}
\begin{table}[!b]
\resizebox{\columnwidth}{!}{%
\begin{tabular}{cccc}
\hline
Method       &  Model                    & \multicolumn{1}{l}{\# Parameters $ (×10^6)$} & Acc \\ \hline
ELE   \cite{madono2020block} & Shakedrop                         & $29.31$  & $83.06$      \\
EtC \cite{chuman2018encryption} & Shakedrop          & $5.35$   & $89.09$     \\
PrivConv \cite{qi2023privacy} & ConvMixer-512/16           & $5.35$   & $92.65$     \\
Ours           &  PreResNet-110                          & $\mathbf{1.70}$    & $\mathbf{95.06}$    \\
\hline
\end{tabular}%
}
\caption{Classification accuracy $(\%)$ on CIFAR-10 dataset for different number of parameters and conventional privacy-preserving image classification methods}
\label{tab:comparison_accuracy}
\end{table}

As presented in Table \ref{tab:comparison_accuracy} our method achieves better state-of-the-art accuracy on private image classification while being highly parameter efficient. Compared to methods like LE and ELE that rely on large models like Shakedrop with over 29 million parameters, primarily due to the need for additional adaptation networks, our approach utilizing a PreResNet-110 architecture attains better accuracy (95.06\%) with over 17x fewer parameters (1.7 million vs 29.31 million). This substantial reduction in parameters demonstrates the ability of our deformable convolution framework to extract meaningful features from protected images without requiring massive overparametrization. Further, our performance surpasses comparable methods like PrivConv built on ConvMixer, showcasing the advantages of leveraging deformable operators over alternative architectures. Crucially, the compact nature of our model allows for privacy-preserving deployment even on resource-constrained platforms. Overall, these results highlight that our proposed technique achieves excellent accuracy and privacy with high efficiency, outperforming existing state-of-the-art methods.

\begin{figure}[!h]
    \centering
    \hspace*{+0.5cm} 
    \includegraphics[width=7cm]{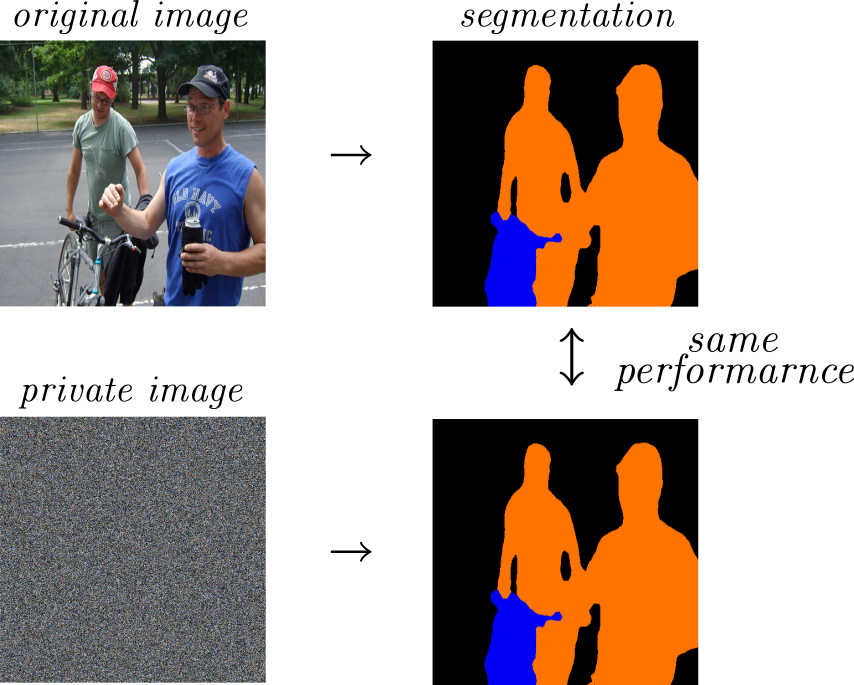}
    \caption{Qualitative results of our proposed framework on the segmentation task. The first row corresponds to the forward pass without privacy, and the second row to the forward pass with privacy.}
    \label{fig:segmentation}
\end{figure}

Finally, to demonstrate the flexibility of our approach for other tasks, we evaluated performance on semantic segmentation using a fully connected network (FCN) \cite{long2015fully} over the PASCAL-VOC dataset, as shown in Figure \ref{fig:segmentation}. Our method can successfully segment objects in the private images; we achieve this by directly loading the trained weights of the FCN model over the original images with the same parameters as the original paper and adapting the layers to our flexible feature learning module, without any fine-tuning on private data. This achieves performance on par with the baseline FCN model on original images, demonstrating the broader applicability of our framework beyond just classification.

\section{conclusion}

This work introduced a novel privacy-preserving deep learning approach using deformable operators to enable secure task learning on private images. Our essential contribution is a framework that applies secret pixel shuffling parametrized by a key during image acquisition; the transformed images are then processed by CNNs enhanced with deformable operators configured with the key. Through extensive experiments on classification and segmentation tasks, we validated that our technique provides robust privacy protection and eliminates the inherent privacy-utility tradeoff and computational costs of prior works. Furthermore, our method avoids cumbersome additions like adaptation networks. Moreover, our approach overcomes state-of-the-art performance while being more parameter-efficient. Our framework provides a foundation for future research, such as developing customized encrypted data transformations and flexible neural network designs for specific tasks.

% #\vfill\pagebreak


\setstretch{0.975}
\bibliographystyle{IEEEbib}
\bibliography{Biblio.bib}

\end{document}